\documentclass[10pt,twocolumn,letterpaper]{article}

\usepackage{iccv}              % To produce the CAMERA-READY version
\usepackage{tablefootnote}
\usepackage{multirow}
\usepackage{booktabs}
\usepackage{graphicx} % For including images
\usepackage{subcaption}
\usepackage{adjustbox}
\usepackage{mwe}

\usepackage{makecell}

\usepackage{comment}
\usepackage{graphicx}

\definecolor{iccvblue}{rgb}{0.21,0.49,0.74}
\usepackage[pagebackref,breaklinks,colorlinks,allcolors=iccvblue]{hyperref}

\def\paperID{13743} % *** Enter the Paper ID here
\def\confName{ICCV}
\def\confYear{2025}

\title{Enhancing Steering Estimation with Semantic-Aware GNNs}

\author{Fouad Makiyeh$^{1}$\thanks{Corresponding Author} \and Huy-Dung Nguyen$^{1}$  \and Patrick Chareyre$^{1}$  \and  Ramin Hasani$^{2}$ \and Marc Blanchon$^{1}$ \; \;\;\;\;\; Daniela Rus$^{2}$ \\
$^{1}$Hybrid Intelligence part of Capgemini Engineering 
        {\tt\small \{first\_name.last\_name\}@capgemini.com}\\
$^{2}$Computer Science and Artiﬁcial Intelligence Lab, Massachusetts Institute of Technology \\
        {\tt\small \{{rhasani,rus}\}@mit.edu}
}

\begin{document}
\maketitle
\begin{abstract}
Steering estimation is a critical task in autonomous driving, traditionally relying on 2D image-based models. In this work, we explore the advantages of incorporating 3D spatial information through hybrid architectures that combine 3D neural network models with recurrent neural networks (RNNs) for temporal modeling, using LiDAR-based point clouds as input. We systematically evaluate four hybrid 3D models, all of which outperform the 2D-only baseline, with the Graph Neural Network (GNN) - RNN model yielding the best results.

To reduce reliance on LiDAR, we leverage a pretrained unified model to estimate depth from monocular images, reconstructing pseudo-3D point clouds. We then adapt the GNN-RNN model, originally designed for LiDAR-based point clouds, to work with these pseudo-3D representations, achieving comparable or even superior performance compared to the LiDAR-based model. Additionally, the unified model provides semantic labels for each point, enabling a more structured scene representation. To further optimize graph construction, we introduce an efficient connectivity strategy where connections are predominantly formed between points of the same semantic class, with only 20\% of inter-class connections retained. This targeted approach reduces graph complexity and computational cost while preserving critical spatial relationships.

Finally, we validate our approach on the KITTI dataset, achieving a 71\% improvement over 2D-only models. Our findings highlight the advantages of 3D spatial information and efficient graph construction for steering estimation, while maintaining the cost-effectiveness of monocular images and avoiding the expense of LiDAR-based systems.

\end{abstract}
\section{Introduction}

\begin{figure}
    \centering
    \includegraphics[scale = 0.5]{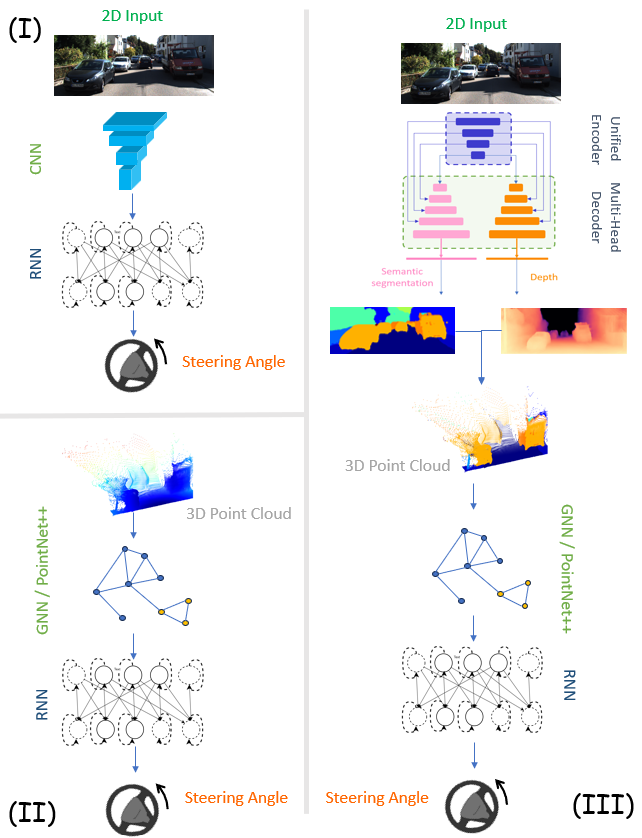}
    \caption{Illustration of Our Approach: From Classical 2D Models to Learned 3D Representations - (I) classical 2D Models, (II) 3D Models with 3D point clouds, and (III) pseudo-point Clouds from monocular images with a learned unified encoder.}
    \label{fig:Approach}
\end{figure}

Autonomous driving systems rely heavily on perception models to estimate steering angles accurately. Traditional approaches, labeled as (I) in Fig.~\ref{fig:Approach}, have predominantly used 2D image-based deep learning models~\cite{bojarski2016end, lechner2020neural}. Although effective, these models lack depth perception and spatial awareness. Recent advances have shown that incorporating 3D information~\cite{yi2022end, chen20203d}, labeled as (II) in Fig.~\ref{fig:Approach}, can significantly enhance performance by capturing road geometry, object distances, and environmental structure, all of which are critical for precise navigation.
%However, relying on LiDAR introduces ........, limiting its scalability in real-world deployments. To address this, we propose an alternative approach that estimates 3D point clouds from monocular images using a learned encoder capable of depth estimation and semantic segmentation. 
LiDAR sensors offer a direct means of obtaining high-precision 3D point clouds~\cite{behley2021towards}, making them a valuable asset for steering estimation~\cite{yi2022end}. They provide accurate depth measurements and enable robust perception in various driving conditions. Nevertheless, LiDAR remains expensive and can only provide a sparse depth map of a scene. In contrast, RGB cameras are significantly more affordable and widely deployed but lack depth perception. To bridge this gap, depth estimation techniques from monocular images have emerged as a promising alternative, allowing to estimate a dense depth map, capable of reconstructing the 3D spatial information from 2D inputs~\cite{bhoi2019monocular, godard2019digging, sun2023dynamo, nguyen2024human}. 
%By integrating estimated depth with traditional visual features, it becomes possible to enhance steering models while maintaining cost efficiency.
In this work, we propose integrating estimated depth with traditional visual features (\eg information about semantic segmentation) to enhance steering models while maintaining cost efficiency.

This study assesses the impact of incorporating three-dimensional data into the estimation of lateral steering, which entails inferring the steering angle of a vehicle as it turns left, turns right, or travels straight ahead.
% In this study, we explore the impact of incorporating 3D information into lateral steering estimation, which involves predicting the vehicle steering angle whatever it turns left, turns right, or moves straight. 
Specifically, we leverage both LiDAR-based point clouds (refer to (II) in Fig.~\ref{fig:Approach}) and semantic point clouds reconstructed from estimated depth maps and semantic maps using the unified encoder along with the original monocular images (refer to (III) in Fig.~\ref{fig:Approach}). Our approach, represented by (II) and (III) in Figure \ref{fig:Approach}, aims to demonstrate the benefits of spatial awareness in improving steering accuracy while considering practical constraints of sensor availability. Using image-derived 3D information, we achieve results comparable to LiDAR-based methods while maintaining the practicality of a camera-only system. Our contributions are:
\begin{itemize}
    \item LiDAR-based steering estimation – We demonstrate that incorporating 3D point clouds improves autonomous vehicules steering  predictions compared to models relying solely on 2D images.
    \item Semantic depth-based pseudo-3D reconstruction – We propose using a unified encoder to estimate depth and semantic maps from 2D images, generating semantic point clouds from a single camera. This technique reduces dependence on LiDAR while preserving critical spatial and contextual information, leading to superior steering performance in a camera-only setup.
    \item Performance evaluation on KITTI – We validate our approach on the KITTI dataset, achieving a 71\% improvement over 2D-only models while leveraging 3D models with 2D image inputs.
\end{itemize}

\section{Related Works}

Few approaches have specifically tackled steering estimation using 3D point clouds, but many have focused on point cloud processing for tasks like classification and segmentation. Given the shared principles of feature extraction across these tasks, we review the relevant literature on 3D neural networks for point cloud processing, alongside existing methods for steering estimation.

\subsection{Point Cloud Feature Learning}

3D Point cloud data poses a unique challenge due to its unordered nature. Traditional voxel-based 3D CNNs process point clouds by discretizing them into volumetric grids~\cite{dai2017scannet}, enabling the application of convolutional operations. However, due to the inherent sparsity of 3D data, these methods suffer from inefficiencies and require high computational resources. To overcome these limitations, Pointnet~\cite{qi2017pointnet} was one of the pioneer methods that directly processed raw point clouds, producing per-point classifications by learning spatial encodings of points. However, PointNet is limited in its ability to capture local structures due to its reliance on a global max-pooling operation. PointNet++~\cite{qi2017pointnet++} extended PointNet by introducing a hierarchical approach that captures local features at multiple scales. This was achieved by recursively applying PointNet on nested partitions of the input point set. 
%The method selects centroids using a farthest point sampling algorithm and groups points into local regions for feature extraction. This multi-scale feature extraction allows for better generalization and adaptability to varying point densities, which makes PointNet++ more effective for complex scenes. Additionally, PointNet++ incorporates a density-adaptive mechanism through two types of layers—Multi-Scale Grouping (MSG) and Multi-Resolution Grouping (MRG)—to intelligently combine features from different scales.
Further developments in point cloud analysis have focused on incorporating contextual information to enhance performance, especially in tasks like semantic segmentation. The 3P-RNN framework~\cite{ye20183d} %proposes an end-to-end approach for unstructured point cloud semantic segmentation. This method 
utilizes an efficient pointwise pyramid pooling module to capture multi-scale context by considering varying neighborhood densities. To address the challenge of long-range dependencies, two-directional hierarchical recurrent neural networks (RNNs) are employed, allowing the model to capture spatial context across large-scale point clouds. 
%This technique provides a robust solution for semantic segmentation by integrating local and long-range spatial features.

Graph Neural Networks (GNNs) have also been employed for processing point cloud data to capture contextual relationships across nodes~\cite{scarselli2008graph}. Each node represents a set of points, and the edges capture spatial relationships. The 3DGNN model~\cite{qi20173d} treated each pixel in the image as a node connected to its nearest neighbors in 3D space and initialized node features using appearance-based descriptors extracted from a 2D CNN applied to RGB images. Through a recurrent update mechanism, nodes iteratively refine their representations based on their own state and messages from neighboring nodes, enabling the network to capture both appearance and 3D geometric information. 
%the proposed GNN-based framework . \review{marc}{missing something}
%This structure enables information propagation across spatially related points while maintaining computational efficiency. Unlike voxel-based methods, this approach directly operates on raw point cloud data, preserving geometric details. By leveraging graph-based message passing, the network iteratively refines feature representations, improving segmentation performance by integrating both 2D image cues and 3D spatial context.
%Recent advances in this area include using superpoint graphs to partition the point cloud into simpler shapes, making the processing more efficient while maintaining high accuracy. These techniques have been extended for 3D object detection and scene understanding, demonstrating the potential of GNNs in improving point cloud processing.
Expanding upon these foundations, Spatial-Temporal Graph Convolutional Networks~\cite{yan2018spatial} have been introduced to model dynamic skeleton-based action sequences. They constructed a graph where nodes represent joints in a skeleton, and edges are formed based on both spatial and temporal relationships. 
%This design enables the network to capture dependencies within a single frame while simultaneously modeling motion dynamics across time. 
A key advantage of this approach is the integration of spatial and temporal information within a unified graph structure.  
VectorNet~\cite{gao2020vectornet} introduced a hierarchical graph representation where individual road components (e.g. lane boundaries, crosswalks) are treated as vectorized polylines. These polylines are embedded into a graph structure, enabling high-order interactions between scene elements and dynamic agents. They constructed local graphs by grouping semantically similar polylines and global graphs using self-attention to exchange spatial and contextual information. The approach integrates perception-based agent dynamics with structured prior knowledge from high-definition maps to predict vehicle trajectories.
%However, a key limitation of this method is its strict connectivity constraints, where only polylines with the same semantics are linked. 

%In contrast, our approach relaxes these constraints by allowing 20\% of connections to remain, even if they do not belong to the same semantic category. This modification enables a richer exchange of spatial dependencies, enhancing feature propagation across heterogeneous map and agent elements.

%Additionally, while VectorNet primarily focuses on trajectory forecasting, our work integrates graph-based representations with sequential models (LSTMs/RNNs) to improve the temporal consistency of steering estimation. By combining LiDAR-based 3D perception with learned scene priors, we enhance the robustness of steering predictions, particularly in challenging urban driving scenarios.
Building on these advancements, we consider both PointNet++ and GNNs in this work to evaluate their effectiveness in feature extraction from 3D point clouds. PointNet++ leverages hierarchical feature learning to capture local and global structures, while GNNs provide a relational understanding of point interactions. By incorporating both approaches, we aim to highlight their complementary strengths and their relevance for steering estimation.
\subsection{Steering Estimation}
Several studies have explored the use of monocular images for end-to-end steering estimation in autonomous driving. 
%These methods leverage deep learning models, primarily convolutional neural networks (CNNs), to extract meaningful features from images and predict steering commands. 
A deep CNN~\cite{bojarski2016end} was trained to map raw pixel data to steering commands. While effective, such purely image-based approaches struggle in complex driving scenarios with varying lighting conditions, occlusions, and dynamic elements.  To enhance robustness, researchers have introduced temporal modeling techniques that integrate past frames using recurrent neural networks (RNNs) or Long Short-Term Memory (LSTM) networks, enabling smoother and more stable steering predictions~\cite{eraqi2017end, xu2017end}.
Furthermore, neuroscience-inspired approaches~\cite{lechner2020neural} have been proposed to improve the interpretability and robustness of autonomous steering systems. 
%Neural Circuit Policies model biological neural circuits to enable structured decision-making with high interpretability. 
These architectures exhibit strong generalization capabilities, making them valuable for safety-critical applications in autonomous driving. Another critical aspect of 2D-based steering estimation is the incorporation of optical flow~\cite{makiyeh2024optical}. They evaluated how different monocular-based modalities, including RGB images, optical flow, and semantic segmentation, contribute to steering estimation performance. 
%Their findings demonstrate that optical flow enhances temporal consistency in steering predictions, reducing abrupt fluctuations caused by frame-by-frame inconsistencies.
On the other hand, LiDAR-based approaches have been explored to map steering decisions directly from point-cloud data. An end-to-end LiDAR-based steering estimation model~\cite{yi2022end} extended the PointNet++ framework to autonomously predict steering commands. This approach formulates the steering problem as a classification task by discretizing steering angles into categories, but only simulation results have been presented.
%The system utilizes K-means, KNN, and weighted sampling to facilitate decision-making. However, .

Unlike purely classification-based approaches, we formulate the steering estimation problem as a regression task, enabling smoother and more precise predictions. Furthermore, our work extends the selected 3D models by integrating sequential modeling through LSTM/NCP to better capture the temporal dependencies of motion trajectories, building on the demonstrated importance of temporal sequencing in 2D models.

\section{Methodology}
\label{sec:methodology}
In this work, we develop a framework for steering estimation in autonomous vehicles that leverages both 3D point clouds and learned 3D representations from monocular images. 
%Our approach consists of three main components:
Our contributions are:
\begin{enumerate}
    \item Architectural Investigation – We systematically evaluate the effectiveness of different neural network architectures for processing 3D point clouds. Specifically, we employ Pointnet++~\cite{qi2017pointnet++} and Graph Neural Networks (GNN)~\cite{kipf2016semi} to extract spatial features, rigorously assessing their respective capacities in capturing geometric structures relevant for steering estimation. 
    Since previous studies~\cite{lechner2020neural, makiyeh2024optical} have highlighted the advantages of temporal modeling, we process these spatial features sequentially using two different recurrent architectures: LSTM and NCP~\cite{lechner2020neural}. 
    % \item Efficient 3D Modeling from Monocular Images for Steering Estimation – To reduce reliance on LiDAR while maintaining 3D scene understanding, we suggest reconstructing 3D point clouds from monocular images using existing depth estimation methods. We achieve this by leveraging a learned unified encoder~\cite{nguyen2024human} for depth estimation, enabling direct comparison between LiDAR-based and image-derived 3D representations. The comparison is conducted using the best model identified in our architectural investigation, where we demonstrate the potential of image-based 3D reconstruction coupled with temporal modeling strategies for improving steering estimation accuracy.
    % \item Towards 3D-Aware Steering Estimation from Monocular Images – We further enhance performance by integrating semantic segmentation from the unified encoder in~\cite{nguyen2024human}. This allows the model to better differentiate between objects of the same class, improving steering estimation in complex driving scenarios.
    \item Towards 3D-Aware Steering Estimation from Monocular Images – To reduce reliance on LiDAR while maintaining 3D scene understanding, we propose reconstructing 3D point clouds from monocular images using a pretrained unified encoder~\cite{nguyen2024human}. By extracting depth maps from monocular images, we enable a direct comparison between LiDAR-based and image-derived 3D representations. We then demonstrate the effectiveness of image-based 3D reconstruction combined with temporal modeling strategies in improving steering estimation accuracy. Additionally, we optimize the 3D model by integrating semantic information from the same unified encoder, reducing model complexity while preserving accuracy.
\end{enumerate}

\subsection{Architectural Investigation}
PointNet++~\cite{qi2017pointnet++} hierarchically learns features by applying three key operations: sampling, grouping, and feature aggregation. Sampling consists of reducing the number of points, a subset \( \mathcal{X} = \{ x_1, x_2, ..., x_M \} \subset \mathcal{P} \) is selected using Farthest Point Sampling (FPS), where \( M \) is the number of sampled points, ensuring uniform coverage of the point cloud. Then, for each sampled point \( x_i \), in grouping they define a local neighborhood \( \mathcal{N}(x_i) \) using a ball query of radius \( r \):
\begin{equation}
    \mathcal{N}(x_i) = \{ x_j \in \mathcal{P} \ | \ \|x_j - x_i\| \leq r \}.
\end{equation}
Each local neighborhood is processed using a shared function \( h_{\theta} \), followed by symmetric aggregation:
\begin{equation}
    f_i = \gamma \left( \{ h_{\theta}(x_j - x_i, f_j) \ | \ x_j \in \mathcal{N}(x_i) \} \right),
\end{equation}
where \( f_j \) is the feature of point \( x_j \), \( h_{\theta} \) is a learnable function (e.g., an MLP) and \( \gamma \) is a symmetric aggregation function, such as max pooling.

On the other hand, the graph network used in this paper is the GCN proposed in~\cite{kipf2016semi}, which updates the node features based on the adjacency structure of the graph. The layer-wise propagation rule for GCN is given in~\cite{kipf2016semi} by:
\begin{equation}
    H^{(l+1)} = \sigma\left( \tilde{D}^{-\frac{1}{2}} \tilde{A} \tilde{D}^{-\frac{1}{2}} H^{(l)} W^{(l)} \right),
\end{equation}
where \( H^{(l)} \) is the node feature matrix at layer \( l \), \( W^{(l)} \) is the trainable weight matrix at layer \( l \), \( \tilde{A} = A + I \) is the adjacency matrix with added self-loops, \( \tilde{D} \) is the diagonal degree matrix of \( \tilde{A} \) and \( \sigma \) is a non-linear activation function (e.g., ReLU).

Moreover, to incorporate temporal dependencies, the spatial features extracted at each time step are processed by a recurrent module. We compare two recurrent neural networks (RNNs): Long Short-Term Memory (LSTM) networks and Neural Circuit Policies (NCP)~\cite{lechner2020neural}. The LSTM is a well-known model, and only the equations for Neural Circuit Policies (NCP) with Liquid Time-Constant Networks (LTC) are presented:
\begin{equation}
\left\{
\begin{aligned}
\tau_i \dot{h}_i &= -h_i + \sum_j w_{ij} \sigma(h_j) + \sum_k w_{ik} x_k + b_i,\\[1mm]
\tau_i &= \exp \left( \theta_i^\top X \right), \; \;  y_i = \sigma(h_i),
\end{aligned}
\right.
\end{equation}
where \( h_i\) is the hidden state of neuron \(i\) at time \(t\), \(\tau_i\)  is the adaptive time constant (learned dynamically), \( w_{ij}\)  are the synaptic weights from neuron \(j\) to neuron  \(j\),
\(\sigma\) is the activation function, \( \theta\) is a learnable parameter, \(X\) is the input feature vector and \(y\) is the NCP output.

To comprehensively evaluate the impact of spatial and temporal modeling choices, we systematically compare four hybrid architectures—(PointNet++ - LSTM), (PointNet++ - NCP), (GNN - LSTM), and (GNN - NCP)—on LiDAR-based 3D representations, providing novel insights into the interplay between spatial abstraction and temporal dependencies in steering estimation. Based on the selected 3D model and RNN, the steering angle $\Theta_t$ is estimated as follows:

\begin{equation}
\label{eqt:steering_lidar}
\Theta_t = \mathcal{R} \Big( h\big(\mathcal{G}(P_t^{(x,y,z)})\big),\dots, h\big(\mathcal{G}(P_{t-\tau}^{(x,y,z)})\big) \Big),
\end{equation}
where $P_t^{(x,y,z)}$ represents the LiDAR point cloud at timestamp $t$, consisting of 3D spatial coordinates, $\mathcal{G}(P_t^{(x,y,z)})$ denotes the transformation of the point cloud into a graph or subset of the point cloud representation via respectively either a GNN or PointNet++, which captures local and global geometric relationships, $h\big(\mathcal{G}(P_t^{(x,y,z)})\big)$ is the extracted hidden state and acting as a compressed latent representation of the scene, and $\mathcal{R}$ is a recurrent module that models the temporal evolution of the vehicule environment over a time-adaptative horizon $\tau$, enabling the system to incorporate a flexible number of past observation. In this setup, the GNN or Pointnet++ modeled as the function $\mathcal{G}$ captures spatial relationships within the point cloud while $\mathcal{R}$ models the temporal evolution of these representation enabling a robust steering angle estimation $\Theta_t$.

\begin{figure}[!htbp]
    \centering
    \includegraphics[scale = 0.22]{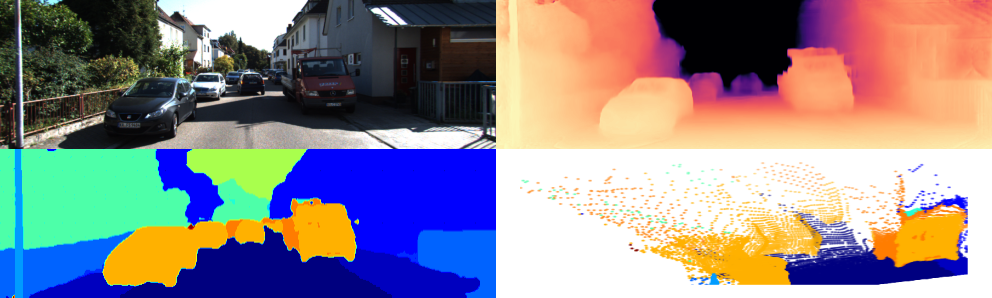}
    \caption{Illustration of original RGB input (top left), estimated dense depth (top right), semantic segmentation prediction (bottom left) and semantic-based point cloud (bottom right).}
    \label{fig:input-illustration}
\end{figure}

\subsection{Towards 3D-Aware Steering Estimation from Monocular Images}
\label{sec:reconstruct_3D}

To alleviate the reliance on LiDAR while maintaining 3D scene understanding, we are interested in reconstructing point clouds from monocular images. To achieve this, we leverage the unified encoder proposed in~\cite{nguyen2024human}, a unified architecture trained on multiple computer vision tasks relevant to urban autonomous driving. The unified encoder is designed to extract dense and generalizable visual representations by jointly learning from various tasks, including depth estimation, pose estimation, 3D scene flow estimation, semantic segmentation, instance segmentation, panoptic segmentation, and motion segmentation. The shared latent space learned by the unified encoder captures rich visual cues, allowing it to provide multiple outputs in a single forward pass efficiently. In our approach, we specifically utilize two outputs generated by the multi head architecture based on the unified encoder: depth and semantic segmentation. These outputs are then used to reconstruct the semantic-based 3D point clouds, which serve as input to our spatial modeling pipeline.

The depth map is back-projected into 3D space using the camera intrinsic matrix \(K\), where each pixel \((u, v)\) in the image is transformed to its corresponding 3D point \((X, Y, Z)\) as follows:
\begin{equation}
\begin{bmatrix}
X , Y , Z
\end{bmatrix}^T
= \mathcal{D}(u, v) \cdot K^{-1}
\begin{bmatrix}
u , v , 1
\end{bmatrix}^T,
\end{equation}
where \(\mathcal{D}(u,v)\) is the estimated depth at pixel \((u, v)\). 
The semantic segmentation is then mapped to the reconstructed point cloud as we know the class for each pixel \((u, v)\), enabling the construction of the semantic-based 3D point cloud. Figure \ref{fig:input-illustration} presents the RGB image with its corresponding depth estimation, semantic segmentation, and reconstructed semantic-based point cloud.

The reconstructed point cloud, whether purely geometric or semantically enriched, is then processed by the spatial modeling module. Formally, the reconstructed point cloud at time \(t\) is denoted as:
\begin{equation}
    P_t^{(x,y,z,c)} = \Pi^{-1}(P_t^{(u,v)}, \mathcal{D}_t(u,v), \lambda \mathcal{S}_t^{(u,v)}),
\end{equation}
where \(P_t^{(x,y,z,c)}\) represents each 3D point with spatial coordinates \((x, y, z)\) and an optional semantic class label \(c\). The projection inverse function \(\Pi^{-1}\) lifts 2D pixel locations \(P_t^{(u,v)}\) into 3D using the estimated depth map \(\mathcal{D}_t(u,v)\), while the semantic segmentation map \(\mathcal{S}_t^{(u,v)}\) assigns class labels to each 3D point, controlled by a factor \(\lambda \in [0, 1]\) referring to the presence of semantic information.

The spatial features extracted from the point cloud at each frame are then processed by a recurrent module to capture temporal dependencies across a sequence of frames. This overall process, combining spatial and temporal reasoning, can be formalized as:
\begin{equation}
\label{eq:withsem}
    \Theta_t = \mathcal{R} \Big( h\big(\mathcal{G}(P_t^{(x,y,z,c)})\big),\dots, h\big(\mathcal{G}(P_{t-\tau}^{(x,y,z,c)})\big) \Big),
\end{equation}
where \(\mathcal{G}\) denotes the spatial feature extraction network (PointNet++ or GNN), \(h\) denotes a feature embedding function, and \(\mathcal{R}\) denotes the temporal modeling module (either LSTM or NCP) responsible for aggregating features over time to estimate the steering angle \(\Theta_t\).

This monocular-based framework allows us to systematically assess the impact of semantic enrichment, spatial modeling strategies, and temporal reasoning mechanisms on steering estimation. By comparing these monocular pipelines to the LiDAR-based approaches, we provide a comprehensive analysis of spatial and temporal modeling across sensing modalities.

\section{Experiments and Results}
\subsection{Dataset}
\label{sec:dataset}
The dataset used in our study for steering estimation is the KITTI raw dataset \cite{geiger2013vision}, which provides multimodal sensor data collected from a vehicle driving in diverse real-world environments. It includes stereo RGB images, LiDAR point clouds, and GPS/IMU measurements across multiple categories, such as city, residential, and highway driving. We use sequences from both the city and residential categories, as they contain varied driving conditions and complex urban scenarios. The dataset also includes vehicle motion data recorded at 100 Hz, including velocity $v$, acceleration, and yaw rate $\dot{\psi}$. These measurements, along with the stereo RGB images and LiDAR point clouds, are key for comparing 2D-based and 3D-based models. To obtain the ground truth steering angle, we use the bicycle model approximation as in~\cite{lechner2020neural}, where the steering angle $\Theta$ (in radian) is computed as:
% \begin{equation} 
%     \Theta = \tan^{-1} \left( \frac{L \cdot \dot{\psi}}{v} \right),
% \end{equation}
$\Theta = \tan^{-1} \left( \frac{L \cdot \dot{\psi}}{v} \right),$
with $L = 2.7m$ represents the wheelbase of the vehicle.

The data is partitioned into sequences. In our case, we use twelve sequences from this dataset for training, two for validation, and one for testing. The test sequence contains over 5,000 frames captured under challenging conditions, including variations in luminosity, multiple road turns, and the presence of multiple vehicles, among others.
\subsection{Results}
\paragraph{2D vs. 3D: Evaluating Spatial Cues for Steering}
We start by comparing the performance of the CNN-NCP model~\cite{lechner2020neural}, a 2D-based approach (refer to (I) in Fig.~\ref{fig:Approach}), with the four aforementioned 3D-based models (refer to (II) in Fig.~\ref{fig:Approach}) introduced in Sec.~\ref{sec:methodology}. Our objective is to prove the impact of spatial information on steering estimation. 
Since the CNN-NCP model emphasizes the importance of temporal context by using a sequence of frames, we maintain this approach across all models. However, to ensure that each model fits within the memory constraints of a single GPU in our setup, the sequence length varies between 8 and 16 frames, depending on the computational demands of each model, which is modelized by $\tau$ in~(\ref{eqt:steering_lidar}) and~(\ref{eq:withsem}). Specifically, we use an NVIDIA RTX A5000 GPU with 24 GB of memory. Furthermore, for 3D-based models, the number of LiDAR points per frame in the KITTI dataset varies significantly, ranging from 75,000 to 160,000. Processing the full point cloud for each frame would exceed memory limits, especially when using longer sequences. To address this, we down-sample the point cloud to either 45,000 or 50,000 points per frame. The choice of down-sampling depends on the complexity of each model, ensuring that all architectures fit within a single GPU while maintaining sufficient spatial resolution for effective learning. Tab.~\ref{Tab:model_parameters} summarizes the sequence length, the number of points in the point cloud, and the parameters of each model, providing a comprehensive comparison of their computational requirements and design choices.

\begin{table}
    \begin{tabular}{ c c c c }
        Model &  RNN & Input size & Sequence size  \\
         \hline
        CNN & NCP  & 192x640 pixels & 16 \\  
        \hline
         Pointnet++ & LSTM & 45k points & 6 \\   
         Pointnet++ & NCP &  45k points & 6  \\   
         GNN & LSTM & 50k points & 8 \\   
         GNN & NCP & 43k points & 8  
    \end{tabular}
    \caption{Comparison of model configurations and parameters used in our experiments.}
    \label{Tab:model_parameters}
\end{table}

Each model is trained for 300 epochs, with early stopping applied when the validation error stopped decreasing.  Given the ground truth steering of the vehicle, $\hat{\Theta}$, the loss function $\mathcal{L}$ to be minimized is as follows:
\begin{align}
    \mathcal{L} &= \frac{\sum_i w(\Theta_i) (\hat{\Theta}_i - \Theta_i)^2}{\sum_i w(\Theta_i)},
\end{align}
where $\Theta_i$ is the steering angle given in~(\ref{eqt:steering_lidar}) as $\Theta_t$ 
%RNN can be either LSTM or NCP, 3D-Model can be either Pointnet++ or GNN, 
and $w(\Theta)$ is a weighting function used in CNN-NCP to emphasize samples containing road turns, while a uniform weight is applied to all points in the 3D models. 
% Fig.~\ref{fig:mse_training} illustrates the evolution of the MSE error for all models throughout the training process, while 
The "Train" column in Tab.~\ref{Tab:MSE_comparisons} presents the average error on the training sequences.
\begin{figure}[http]
    \centering
    \begin{subfigure}[t]{0.23\textwidth}  % Set width to 48% of the line
        \centering
        \includegraphics[scale = 0.5]{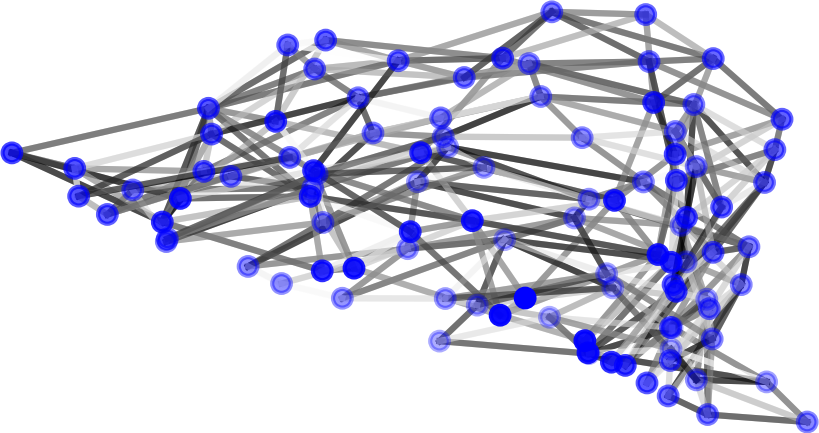}
    \end{subfigure}
    \begin{subfigure}[t]{0.23\textwidth}  % Set width to 48% of the line
        \centering
        \includegraphics[scale = 0.5]{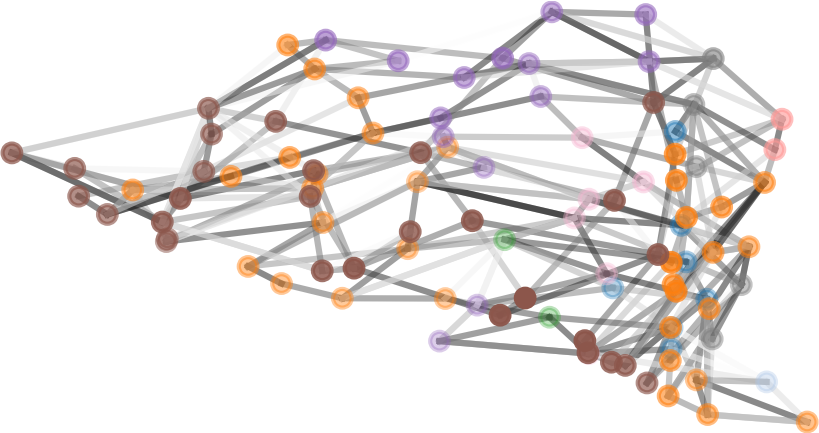}
    \end{subfigure}
    \caption{Illustration of graph reduction and semantic graphs. Left is initial graph from down-sampled point cloud, right is semantic-aware graph where color nodes represent classes. 
    %Edges are colored according to the inter-node weight. 
    Filtering enable reduction of edges from $600$ on the left to $296$ on the right. }
    \label{fig:graph_filter}
\end{figure}
The comparison of the mean square error (MSE) between 2D and 3D models on the test sequence, as summarized in Tab.~\ref{Tab:MSE_comparisons}, highlights the importance of spatial information in steering estimation. For instance, by comparing the rows within the "Test" column for groups I and II, all 3D models achieve superior performance compared to the 2D model, as the MSE obtained for all models in group II is lower than that of the CNN-NCP from group I, which is $0.2676$. This validates our first contribution on improving steering estimation on Kitti dataset, as 3D models extract richer geometric features that enhance scene understanding. 
%All models trained on 3D point clouds consistently outperform the 2D model in steering estimation 

This performance is due to several key advantages. First, 3D point clouds provide explicit depth information, allowing the model to estimate distances to obstacles and lane boundaries more accurately, whereas 2D images only offer implicit depth cues that may be ambiguous in monocular settings. This difference is particularly crucial when handling turns, elevation changes, and multi-lane roads, where perspective distortions in 2D images can lead to incorrect predictions. Second, 3D representations remain robust under varying lighting conditions, such as shadows, glare, which often degrade image-based models. Since point clouds capture geometric structure rather than appearance-based features, they are less affected by visual artifacts. Additionally, occlusions are better handled in 3D because a single 2D image may lack the necessary context due to object occlusions or limited field of view. Finally, 3D models better encode spatial relationships between objects, enabling a more structured understanding of the scene, which improves generalization across different driving conditions.
In summary, depth information contributes to more accurate trajectory predictions, particularly in complex driving conditions where occlusions and perspective distortions pose challenges for purely image-based approaches. 

Within the 3D-based methods (Group II), models leveraging GNNs outperform those based on PointNet++ due to their ability to explicitly capture 3D spatial proximity between points. Unlike PointNet++, which processes point clouds hierarchically using local feature aggregation, GNNs construct a graph representation where points (or clusters of points) serve as nodes, and edges encode meaningful spatial connections. 
%This structure enables GNNs to model both local and global geometric dependencies more effectively. 
Furthermore, GNNs dynamically update edge weights based on learned features, allowing the model to focus on critical regions, such as lane boundaries and obstacles, rather than treating all points equally. Moreover, GNNs inherently facilitate information propagation across neighboring points, leading to smoother and more coherent feature extraction, while PointNet++ relies on predefined grouping strategies that may not optimally capture long-range dependencies.

\begin{table}
    \begin{tabular}{ c c c c c c }
        % & \multicolumn{3}{c}{MSE ($radian^{2}$)} \\
        Group &  Approach & $\mathcal{D}$ & $\mathcal{S}$ & Train & Test \\
        \hline
        I & CNN-NCP & & & 0.108 \tablefootnote{For the CNN-NCP training, we use weighted mse as suggested in~\cite{lechner2020neural}}  & 0.267 \\  
        \hline
        \multirow{4}{*}{II} & GNN-LSTM & & & 0.033 & 0.130 \\   
        & \underline{GNN-NCP} & & & \underline{0.046} & \underline{0.132} \\   
        & Pointnet++-LSTM & & & \textbf{0.032} & 0.182 \\   
        & \underline{Pointnet++-NCP} & & & \underline{0.053} & \underline{0.240} \\   
        \hline
        \multirow{2}{*}{III} & \underline{$\mathcal{D}$o-GNN-NCP}  & \checkmark & & \underline{0.038} & \underline{0.081} \\  
         & \underline{$\mathcal{S}$a-GNN-NCP} & \checkmark &   \checkmark &       \underline{0.037}                  &                \underline{\textbf{0.077}}
    \end{tabular}
    \caption{MSE ($radian^{2}$) for training and test values across different models, categorized by 2D-based (I), LiDAR-based (II), and reconstructed point cloud-based models (III). $\mathcal{D}$o refers to Depth-Only or Point cloud projected from depth and $\mathcal{S}$a refers to Semantic-Aware Point Cloud. \textbf{bold} is best and \underline{underline} is \textit{ours}.  }
    \label{Tab:MSE_comparisons}
\end{table}

% \begin{figure}
%     \centering
%     \includegraphics[scale = 0.3]{Images/Error_variation.png}
%     \caption{Evolution of MSE for different 3D-based models during the experiments .}
%     \label{fig:mse_training}
% \end{figure}

\paragraph{Comparison of LiDAR-Based and Image-Derived 3D Point Clouds }

In this section, we compare the performance of steering estimation models using LiDAR-based point clouds with those using 3D point clouds reconstructed from monocular images as explained in Sec.~\ref{sec:reconstruct_3D}. 
Specifically, the GNN-based models achieved the best performance among the 3D approaches.
Since the performance of GNN-NCP and GNN-LSTM is comparable, we choose GNN-NCP for this comparison, as it offers better interpretability due to its biologically inspired liquid neuron dynamics, as demonstrated in~\cite{lechner2020neural}.

We begin by reconstructing the 3D point clouds using monocular images along with the estimated depths. The first comparison is conducted by analyzing the rows corresponding to GNN-NCP from Group II and proposed-GNN-NCP-only-depth from Group III, focusing on the "Test" column in Tab.~\ref{Tab:MSE_comparisons}.
We can notice that steering estimation is significantly more precise when using reconstructed point clouds. This improvement can be attributed to several key factors. Despite LiDAR superior depth accuracy, the reconstructed point clouds offer advantages that enhance their performance in steering estimation. Monocular depth estimation models learn from large datasets of real-world driving scenes, they inherently encode information about object boundaries and surface continuity. This results in more spatially coherent dense depth maps, reducing sparsity and noise. This density can be beneficial when extracting features for steering estimation, as the model receives a more continuous spatial representation of the environment. In contrast, LiDAR captures a sparse set of points, especially at longer distances. Furthermore, monocular depth estimation integrates contextual cues from RGB images, such as texture and shading, which can provide additional depth refinement beyond geometric measurements alone. 
%As a result, while LiDAR remains a valuable sensor for 3D scene understanding, monocular image-based 3D reconstruction presents a promising alternative for steering estimation, particularly when sensor cost and deployment constraints are considered.

    \begin{figure*}[htbp]
    \centering
    \begin{adjustbox}{max width=\textwidth}
    \begin{subfigure}[b]{0.249\textwidth}
        \centering
        \includegraphics[width = \linewidth]{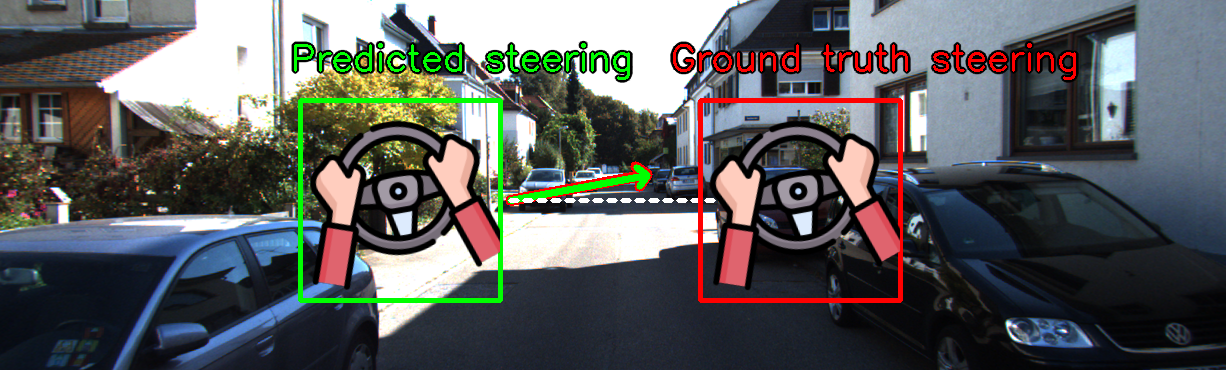}
        \caption{Minor left turn}
        \label{fig:sfig1}
    \end{subfigure}
    \begin{subfigure}[b]{0.249\textwidth}
        \centering
        \includegraphics[width = \linewidth]{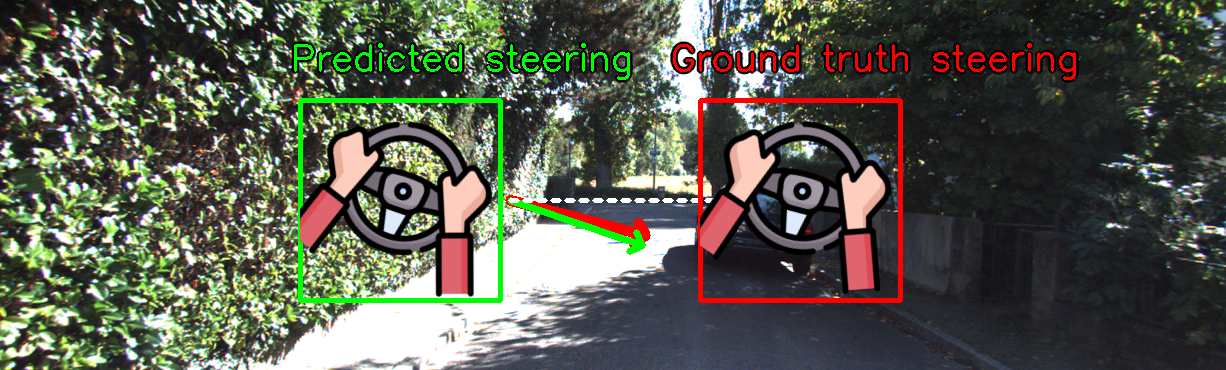}
        \caption{Slight right turn}
        \label{fig:sfig2}
    \end{subfigure}
    \begin{subfigure}[b]{0.249\textwidth}
        \centering
        \includegraphics[width = \linewidth]{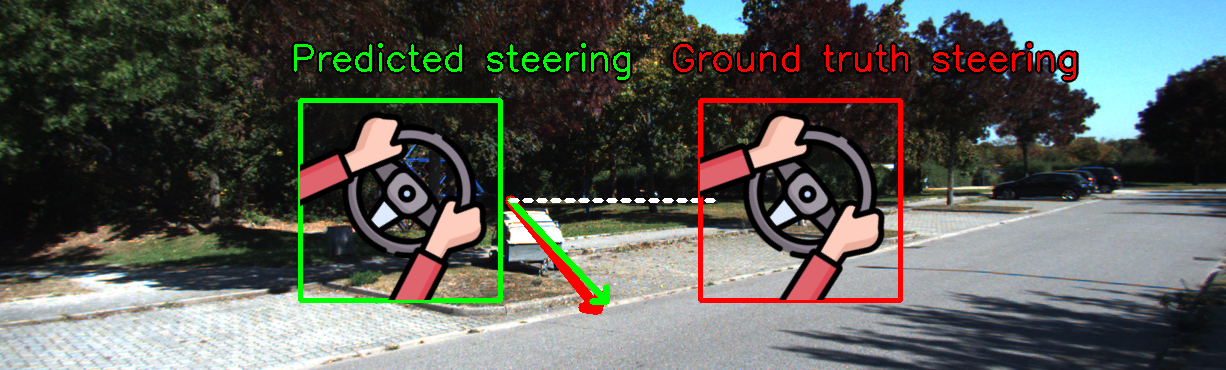}
        \caption{Sharp right turn}
        \label{fig:sfig3}
    \end{subfigure}
    \begin{subfigure}[b]{0.249\textwidth}
        \centering
        \includegraphics[width = \linewidth]{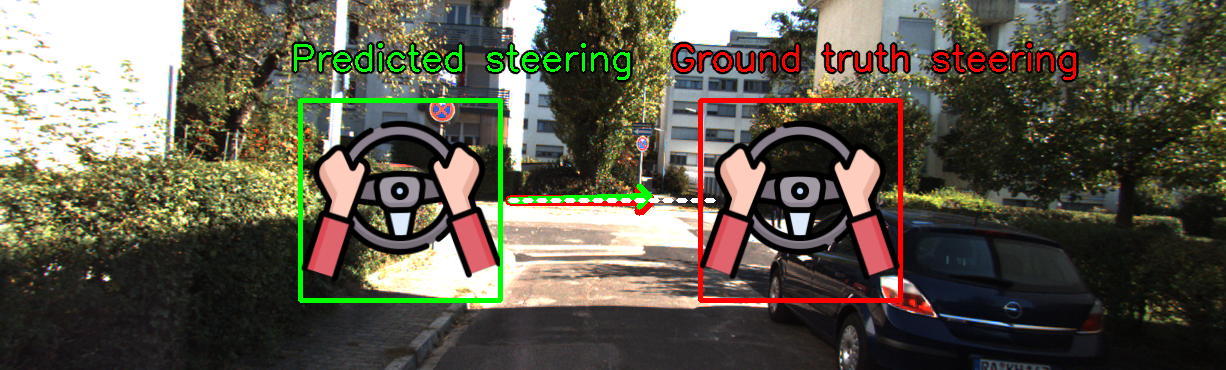}
        \caption{Driving straight}
        \label{fig:sfig4}
    \end{subfigure}
    \end{adjustbox}
    
    \caption{Visualization of different driving scenarios. Icons and arrows are \textcolor{red}{red} for ground truth and \textcolor{green}{green} for prediction.}
    \label{fig:steering_image_plot}
\end{figure*}

% Furthermore, we optimize the graph by reducing edge connectivity, making the system less computationally expensive by incorporating semantic information. When points belong to the same class, the edges are retained. However, when points do not belong to the same class, only 20\% of the edges are retained. 
Furthermore, we optimize the GNN by reducing edge connectivity, making the system less computationally expensive by incorporating semantic information. 
The semantic map used for this process is obtained from the same unified encoder used previously for depth estimation, which means that no additional computational complexity is introduced. 
Specifically, when two points in the graph belong to the same class, the edge connecting them is fully retained, ensuring that semantically meaningful structures remain intact. However, when two points belong to different classes, only 20\% of the edges between them are preserved. For illustrative purposes, we down-sampled the point cloud of a random frame from the test sequence to 100 points, as shown in Fig.~\ref{fig:graph_filter}. The graph constructed before optimization is shown on the left and the graph after optimization is shown on the right in Fig.~\ref{fig:graph_filter}.
This reduction in graph complexity offers several advantages. First, it reduces noise by limiting connections between unrelated objects, such as pedestrians or background structures, which could interfere with steering estimation. 
Second, it preserves local structure by maintaining key spatial relationships within each semantic class, ensuring that essential geometric features, such as road surfaces and lane markings, remain well-connected. 
Third, it improves generalization by removing unnecessary edges that could lead to overfitting, allowing the model to focus on meaningful patterns rather than irrelevant details. 
Fourth, the computational efficiency gained by reducing the number of edges leads to more stable training dynamics, preventing noisy updates that could degrade performance. 
%Finally, this approach enhances semantic relevance by prioritizing connections within critical regions for steering, ensuring that the model focuses on structurally important areas rather than incorporating misleading relationships. 
% Despite the reduced graph complexity, the model achieves comparable or even slightly improved steering accuracy, highlighting the benefits of structured graph simplification.
%This approach enables a significant reduction in graph computation while still preserving crucial spatial relationships, ultimately leading to improved generalization. 
The effectiveness of our optimized method is evident in the results presented in group III in Tab.~\ref{Tab:MSE_comparisons}. Specifically, the MSE of "$\mathcal{S}$a-GNN-NCP" is consistently lower than that of "$\mathcal{D}$o-GNN-NCP", demonstrating that our strategy of leveraging semantic information enhances accuracy while simultaneously reducing complexity. 
%This validates our core motivation: by filtering graph edges based on semantic consistency, we not only remove redundant or misleading connections but also reinforce meaningful structural dependencies that are more relevant to steering estimation. 
Moreover, since the semantic segmentation is obtained from the same unified encoder used for depth estimation, our approach does not introduce additional computational overhead. 
These results highlight a key contribution of our work—showing that intelligent graph pruning, guided by semantic information, is an effective strategy for optimizing graph-based neural networks in autonomous driving applications.

In summary, these observations validate our second contribution, highlighting the effectiveness of reconstructed point clouds in steering estimation. By demonstrating the advantages of monocular depth estimation over LiDAR in specific scenarios, we emphasize its potential as a competitive alternative, especially where sensor cost and deployment constraints are significant. 
Additionally, our results validate an additive refinement to our second contribution, showcasing how leveraging semantic information for graph pruning enhances accuracy while reducing computational complexity. 
By intelligently filtering graph edges based on semantic consistency, we reinforce meaningful structural dependencies crucial for steering estimation, without adding extra computational overhead. 
Compared to 2D models, our approach achieves a $71\%$ improvement, as demonstrated by the MSE obtained with 2D models $0.2676$ versus our final model $0.0771$.
This highlights the effectiveness of our proposed strategy in optimizing graph-based neural networks for autonomous driving.

\paragraph{Illustrating estimated steering on 2D Images}
In this section, we present in~Fig.~\ref{fig:steering_image_plot} qualitative results by visualizing the estimated steering angles using samples from the test dataset. Since directly displaying the results on 3D point clouds, which often contain over 100k points, would be complex and cluttered, we instead overlay the predictions on the corresponding 2D images from the KITTI dataset. It is important to emphasize that these images are used only for visualization, while both training and testing are conducted entirely on 3D point clouds. The KITTI dataset provides synchronized 3D LiDAR point clouds along with RGB images, allowing for a clear and interpretable representation of the model predictions.

To illustrate the steering direction, we display a steering wheel icon\footnote{You can download the icon from the following \href{https://www.flaticon.com/free-icons/driving}{link}} that is rotated according to the predicted and actual steering angles. To clearly differentiate between the two, we enclose the predicted steering wheel in a green box, while the ground truth is shown in a red box. This allows for a direct visual comparison of how well the model captures steering dynamics.  In addition, we added two arrows (green for the estimated steering angle and red for the ground truth) to enhance the visualization of the steering angle accuracy. Additionally, the dashed white line represents the reference axis at 0 radian. To further illustrate the performance of our model, we provide different representative cases: (a) a minor left turn, (b) a slight right turn, (c) a sharp right turn, and (d) driving straight. These cases highlight the model ability to handle various driving scenarios, including sharp turns and minor steering adjustments. 
%Additionally, we overlay the steering error directly on the images, providing a numerical indication of the difference between the predicted and actual steering angles.  
We can observe that the steering icon is rotated very closely when using the predicted steering angle compared to the one using the ground truth for these different cases. Additionally, we can see that the green and red arrows are rotated with respect to the dashed lines by almost the same angle. This demonstrates the accuracy and precision of our model on a dataset that includes different cars in a single scene, with luminosity changes, which can be a major limitation for 2D models. Since this is done on a test dataset, it further highlights the generalization success of the proposed model.
%The corresponding images are extracted from the test dataset and serve as qualitative evidence of the model effectiveness in steering estimation.

\begin{figure}
    \centering
    \includegraphics[scale = 0.24]{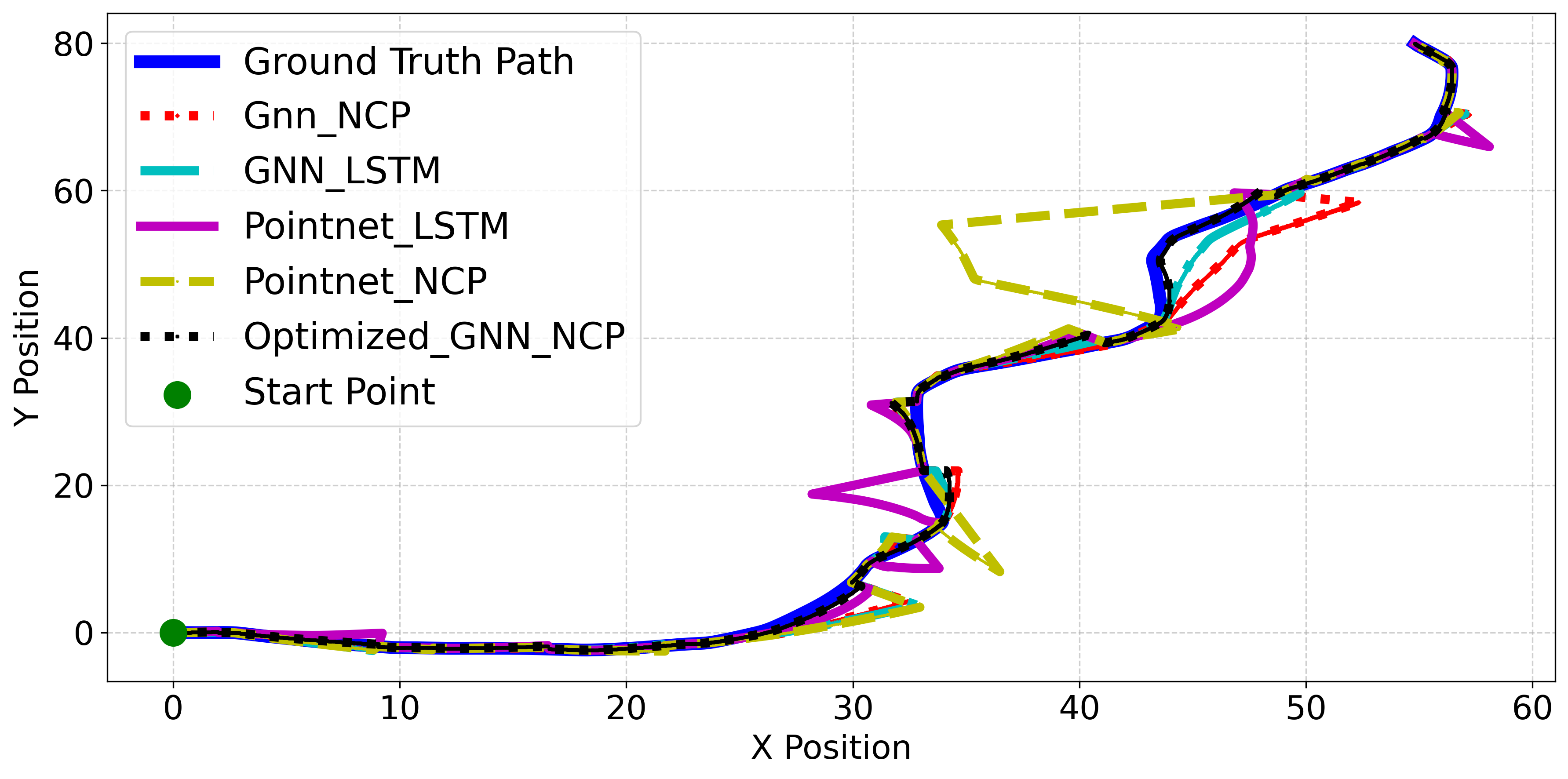}
    \caption{Predicted vehicle trajectories using steering estimation from different 3D models.}
    \label{fig:path_estimation}
\end{figure}

\paragraph{Additional Application: Path Estimation from Steering Angle }
Beyond directly estimating the steering angle, our model can reconstruct the vehicle traveled path. 
Given an initial position (green dot in Fig.~\ref{fig:path_estimation}), the trajectory is obtained by integrating the predicted steering angles with the vehicle velocity provided by the IMU sensor (refer to Sec.~\ref{sec:dataset} for more information). This enables a visual representation of the path the vehicle would have followed based on the model predictions.

A key challenge in this process is the accumulation of errors over time. Since each predicted steering angle determines the next position, even small errors can propagate, resulting in a significant drift between the predicted and actual trajectories. Because correcting these deviations is beyond the scope of this work, errors continue to accumulate, even if subsequent steering predictions are accurate. This issue is particularly critical at decision-making points, such as intersections, where the model must determine whether to turn left, right, or proceed straight. In these scenarios, mispredictions exacerbate trajectory drift. 
To mitigate this effect, we periodically reset the predicted path to align with the ground truth at critical decision points. However, once the vehicle has initiated a turn or continued straight, the model reliably predicts the steering angle, as the dynamics of the motion provide strong contextual cues. 
%Thus, the primary source of error is not in executing turns but in determining in which direction to turn. 
Although our model effectively estimates steering for continuous motion, it does not perform high-level decision making, which remains an open challenge beyond the scope of this study.

To assess trajectory reconstruction accuracy, Fig.~\ref{fig:path_estimation} presents the paths generated using steering angles estimated by different 3D models, along with the ground truth trajectory. To ensure a fair comparison, all models undergo the same reinitialization process at intersection points. As before, the illustrated results are obtained from the test dataset. The trajectory is reconstructed by integrating the predicted steering angles with the vehicle velocity from the IMU and the time step $\Delta t$, following:
% \begin{align} 
% x_{t+1} &= x_t + v_t \cos({\Theta}_t) \Delta t \\
% y_{t+1} &= y_t + v_t \sin({\Theta}_t) \Delta t 
% \end{align}
\begin{equation*}
\begin{bmatrix}
x_{t+1}, y_{t+1} 
\end{bmatrix}^T
= \begin{bmatrix}
x_{t}, y_{t} 
\end{bmatrix}^T 
+ v_t \Delta t
\begin{bmatrix}
\cos({\Theta}_t), \sin({\Theta}_t) 
\end{bmatrix}^T
\end{equation*}
Furthermore, the MSE values presented in Tab.~\ref{Tab:MSE_comparisons} directly correlate with the precision of trajectory reconstruction. For instance, the most divergent path from the expected one is the yellow and purple plots, which correspond to the PointNet++-based model, which has the highest MSE among the 3D models. In particular, our final proposed model, group (III), which has the lowest MSE, demonstrates a trajectory reconstruction that closely aligns with the actual vehicle path, underscoring its improved steering prediction accuracy. This result underscores once again the strength of our approach, this time in improving trajectory estimation.

\section{Conclusion}
In this paper, we introduced a novel approach to steering estimation for autonomous vehicles leveraging 3D information. We conducted an extensive comparison of 3D hybrid models using LiDAR-based point clouds, evaluating various architectures for steering estimation. Our results revealed that graph-based networks (GNNs) consistently achieved the best performance, effectively capturing spatial relationships within the 3D data.

Although LiDAR-based point clouds are widely used, we demonstrated that reconstructed pseudo point clouds from monocular images—generated via a learned unified encoder—achieved even better performance. This finding underscores the potential of monocular-based perception as a cost-effective yet highly accurate alternative to LiDAR. Additionally, we optimized the GNN architecture using semantic maps extracted from the same unified encoder, further refining spatial reasoning within the model.
Our proposed approach resulted in a 71\% improvement over 2D-only models on real-world urban driving data from the KITTI dataset. Beyond steering estimation, we showcased its versatility in trajectory path prediction, validating its robustness in various driving conditions, including straight roads, minor turns, and sharp curves.

By demonstrating that monocular-based pseudo point clouds surpass LiDAR-based representations, our study challenges conventional assumptions in autonomous vehicle perception. This breakthrough paves the way for cost-efficient and high-performance 3D perception, opening new possibilities for monocular-driven autonomous navigation.

{
    \small
    \bibliographystyle{ieeenat_fullname}
    \bibliography{main}
}

\end{document}